\documentclass[11pt]{article}

\usepackage{tikz}
\usepackage{enumerate}
\usepackage{enumitem}
\usepackage{blindtext}
\usepackage[T1]{fontenc}
\usepackage{microtype}
\usepackage{graphicx}
\usepackage{booktabs} 
\usepackage{algorithmic}
\usepackage{algorithm}
\usepackage{times}
\usepackage{epsfig}
\usepackage{amsmath}
\usepackage{amssymb}
\usepackage[font=small,labelfont=bf]{caption}

\usepackage{array}
\newcolumntype{P}[1]{>{\hspace{0pt}}p{#1}}

\usepackage{comment}
\usepackage[export]{adjustbox}
\usepackage{rotating}
\usepackage{csvsimple,longtable,booktabs}
\usepackage{tabularx}
\usepackage[section]{placeins}

\usepackage{hyperref}
\usepackage{authblk}

\begin{document}

\title{Graph Minors Meet Machine Learning: the Power of Obstructions}

\author[1]{Faisal N. Abu-Khzam}
\author[2]{Mohamed Mahmoud Abd El-Wahab}
\author[3]{Noureldin Yosri}

\affil[1]{Department of Computer Science and Mathematics, Lebanese American University, Lebanon}

\affil[2]{School of Engineering, Information Technology \& Environment,
Charles Darwin University, Australia}
\affil[3]{College of Computing and Information Technology, Arab Academy of Science and Technology, Egypt}

\date{}
\maketitle

\begin{abstract}
    Computational intractability has for decades motivated the development of a plethora of methodologies that mainly aimed at a quality-time trade-off. The use of Machine Learning techniques has finally emerged as one of the possible tools to obtain approximate solutions to ${\cal NP}$-hard combinatorial optimization problems. In a recent article, Dai et al. introduced a method for computing such approximate solutions for instances of the Vertex Cover problem. In this paper we consider the effectiveness of selecting a proper training strategy by considering special problem instances called {\em obstructions} that we believe carry some intrinsic properties of the problem itself. Capitalizing on the recent work of Dai et al. on the Vertex Cover problem, 
    and using the same case study as well as 19 other problem instances, we show the utility of using obstructions for training neural networks. Experiments show that training with obstructions results in a huge reduction in number of iterations needed for convergence, thus gaining a substantial reduction in the time needed for training the model.
\end{abstract}

\section{Introduction}

Let $G$ be a finite undirected graph with possible loops and multiple edges. A graph $H$ is said to be a minor of $G$, or $H \leq_m G$, if a graph isomorphic to $H$ can be obtained from a subgraph of $G$ by contracting zero or more edges. Here an edge $uv$ is contracted by replacing $u$ and $v$ with a vertex $w$ whose neighborhood is the union of the neighborhood of $u$ and $v$.
As such, $\leq_m$ defines a partial ordering on the set of finite graphs known as the minor order.

Let $F$ be a family of finite graphs that is closed under the minor order. 
The obstruction set of $F$ is defined to be the set of graphs in the complement of $F$ that are minimal in the minor order. That is $X$ is an obstruction of $F$ if $X\notin F$ and every minor of $X$ is in $F$.
A typical classical example of a minor-closed family is the set of planar graphs. In this case, and according to Wagner's theorem \cite{Wagner1937}, the set of obstructions consists of only two graphs: the complete graph on five vertices and the balanced complete bipartite graph on six vertices.

Thanks to the celebrated Graph Minor Theorem, which states that finite graphs are well quasi ordered by the minor order relation \cite{ROBERTSON2004325}, we know 
the obstruction set of a minor-closed family is finite. This implicitly leads to polynomial-time membership tests for such families, though non-constructively according to the work of Fellows and Langston \cite{FellowsLangston88}, which paved the way for the emergence of {\em parameterized complexity} theory \cite{DF99}. The main idea being that a decision problem whose family of YES-instances is minor-closed would be efficiently solvable in polynomial time when some input parameter is fixed. A well studied notable example is the classical {\em Vertex Cover} problem. Unfortunately, there is no general algorithm for constructing sets of vertex-cover obstructions \cite{Dinneen07}. In this paper we address the optimization version of the problem and show that computing a potentially large subset of such obstructions can be used to improve vertex cover approximation via a proper use of deep learning.

Using deep learning to solve graph theoretic problems is considered hard in general, even if approximate solutions are sought. This can be attributed to the difficulty of representing/organizing graphs in a grid-like structure, and deal with various forms of graphs: weighted/unweighted, directed/undirected, etc.... Moreover, the nature of a required solution varies from one problem to another (vertex/edge classification, graph classification, etc.)
To tackle these challenges, different neural network architectures have been proposed including Graph Neural Networks (GNNs) \cite{Gori2005ANM,FrasconiAdaptiveProcessing,ScarselliGNN}, 
Graph Convolution Networks (GCNs) \cite{BrunaGCN}, and Graph Autoencoders (GAEs) \cite{KipfGAE}.

An in depth discussion for problems to apply deep learning to graphs as well as description and comparison of various models was done by Zhang et al. in \cite{Zhang2018}.
To tackle combinatorial optimization problems on graphs, Vinyals et al. introduced the Pointer Net (Ptr-Net) architecture \cite{NIPS2015_5866}, while Bello et al. proposed a reinforement learning approach to combinatorial problems \cite{Bello2016}. 
In this work we build on the work of Dai et al. in \cite{Dai16} and  \cite{Dai18}, namely using the Structure2Vec model and the reinforcement learning formulation of Bello et al. in \cite{Bello2016}.

\section{Preliminaries}

Throughout the paper we use common graph theoretic terminology. For a graph $G$, we use $V(G)$ and $E(G)$ to denote the sets of vertices and edges of $G$, respectively. For a vertex $v$ of $G$, we denote by $N(v)$ the neighborhood of $v$, i.e., the set of vertices adjacent to $v$ in $G$, and by $E(v)$ the set of all edges that are incident on $v$. 

A vertex cover of a graph $G$ is a subset of the vertex set of $G$ whose deletion results in an edge-less graph. In other words, every edge of $G$ has at least one endpoint in $C$. The corresponding optimization problem, which seeks a minimum-cardinality vertex cover in a given graph, is one of the most studied classical ${\cal NP}$-hard problems \cite{Karp1972}. 

An effective greedy approach for Vertex Cover consists of successively selecting a vertex of maximum degree and placing it in the solution, until the graph is edge-less. A pseudocode is given below.

\begin{algorithm}
\caption{}
\label{greedy1}
\begin{algorithmic}
\STATE $\textit{C} \gets \phi$
\WHILE {$\textit{there is at least one uncovered edge}$}
\STATE $u \gets argmax_{v \notin \textit{C}} \|\{vt | t \notin C\}\|$
\STATE $\textit{C} \gets \textit{C} \cup \{u\}$
\ENDWHILE
\STATE $\textbf{Return} \ \  C$
\end{algorithmic}
\end{algorithm}

The above method does not guarantee a constant-factor approximation, but it is known to perform well in practice. In fact the problem is known to be ${\cal APX}$-complete with a best-known factor-two polynomial-time approxiamtion algorithm, attributed to F. Gavril in \cite{GJ79} but also known to be independently discovered by M. Yannakakis. This was shown to be best possible modulo the unique games conjecture \cite{KHOT2008335}. A pseudocode of this factor-two approximation algorithm follows. In the next section, we present a reinforced machine learning approach that can compete with known approximation algorithms.

\begin{algorithm}
\caption{}\label{greedy2}
\begin{algorithmic}
\STATE $\textit{C} \gets \phi$
\WHILE {$\textit{there is at least one uncovered edge}$}
\STATE $uv \gets \textit{any uncovered edges}$
\STATE $\textit{C} \gets \textit{C} \cup \{u,v\}$
\ENDWHILE
\STATE $\textbf{Return} \ \  C$
\end{algorithmic}
\end{algorithm}

For any (fixed) integer $k$, the family $F_k$ of graphs of minimum vertex cover bounded above by $k$ is minor-closed. This is simply due to the fact that neither edge contraction nor vertex/edge deletion can increase the minimum vertex cover size. Therefore we know the set of obstructions for $F_k$ is finite. We shall denote this set by $\mathcal{O}b(k)$. We also use $vc(G)$ to refer to a minimum-size vertex cover of $G$.
We should note that many combinatorial graph problems satisfy the same closure property (under taking minors) when fixing some parameter that is often the solution size. Notable examples include Treewidth, Pathwidth and Feedback Vertex Set. We shall refer to such problems as being {\em minor-closed} in the rest of this paper.

We denote by $C_n$ the cycle on $n$ vertices, and by $K_n$ the complete graph (or clique) on $n$ vertices.
Observe that a graph $G$ that has $C_{2k+1}$ as minor must have a minimum vertex cover of size strictly greater than $k$. The same applies if $K_{k+2} \leq_m G$.
Thus the obstruction set of a minimum vertex cover of size $k$ contains $C_{2k+1}$ and $K_{k+2}$. 
Unfortunately, it was shown in \cite{Cattell94} and \cite{Dinneen07} that the size of the obstruction set for $k$-Vertex Cover grows exponentially with $k$. 
In general, there is no constructive approach to generate all the connected obstructions of $k$-Vertex Cover from the obstructions for $(k-1)$-Vertex Cover \cite{Dinneen07}. However, a large percentage of such obstructions can be obtained, especially when $k$ is small enough, as we shall see in the next section.

\section{Deep Learning for Minor-Closed Combinatorial Problems: the case of Vertex Cover}

In \cite{Dai18}, Dai et al. used the structure2vec model of \cite{Dai16} to solve approximately a number of combinatorial optimization problems including Minimum Vertex Cover. The main setup formulates the problem as a reinforcement learning problem as in \cite{Bello2016} 
applied to an Erdos-Renyi model for graph generation and to the meme-tracker dataset \footnote{http://www.memetracker.org/}.
As mentioned earlier, we build on this recent work and we further suggest the use of Vertex Cover obstructions of sizes ranging from 1 to 10 
for training, instead of using random subgraphs of the target graph as done in \cite{Dai18}.
The rational behind this approach stems from the idea that when training a network on a specific family of graphs that share the same statistics, the network should be able to learn those statistics and to maximize a function mapping  the given choices for a move (e.g. the next vertex to add to the vertex cover) to a cost representing the true cost/gain of using that vertex.
For example, in the Erdos-Renyi model for graph generation any edge may exist with probability \textit{p}, and so all graphs on $n$ vertices and $m$ edges exist with equal probability $p^m (1-p)^{{n \choose 2} - m}$. These graphs share some common statistics, such as expected degree and cluster distribution; so given a large sample of these graphs, a large enough network should be able to effectively learn those statistics.

The Structure2Vec approach introduced by Dai et al. \cite{Dai16} is a new model that computes a \textit{p} dimensional feature embedding $\mu_v$ for each vertex $v \in V$.
It does that by initializing $\mu^{(0)}_v = 0$ and then update the embedding synchronously at each iteration as:
\[\mu^{(t+1)}_v = F(x_v, \{\mu^{(t)}\}_{u\in\mathcal{N}(v)}, \{w(v, u)\}_{u\in\mathcal{N}(v)}; \Theta)\]

\noindent
Where $x_v$ is a vertex specific tag (in our case whether \textit{v} is in the current vertex cover or not),
$w(v,u)$ is the weight of edge v=$vu$, which is set to 1 in our experiments, and \textit{F} is a non linear function and $\Theta$ is the parameter of the network.

We recall that in a reinforcement learning setting for a certain problem there are typically four main elements (among other things): the problem {\em state}; a set of possible {\em actions}; the model that interprets the state and decides what action to take (referred to by the {\em agent}); and a {\em reward function} that takes a (state, action) pair and returns the reward for taking that action.
The goal of the agent is to maximize the summation of rewards it receives, in our case the setting is:
\begin{itemize}
    \item \textbf{State} $\sum_{v\in V} \mu_v$ 
    \item \textbf{Actions} the choices for which vertex to add to the cover $v \notin V \wedge v \in V$  
    \item \textbf{Agent} the model
    \item \textbf{Reward function} the reward for each move the constant -1
\end{itemize}

Training the model happens in steps. The first is to randomly initialize the weights of the network, then sample a batch of data to serve as a cross validation data, then we sample batches of data for training, after every epoch we evaluate the current model against the validation data and finally pick the best performing model on the validation date; while training we also evaluate the models on some popular graph data sets to monitor how the performance change (however this does not change training in anyway since we only visualize the performance on the test data but do not decide on it). A few samples of those visualizations are presented in Figures \ref{fig:1}, \ref{fig:2}, \ref{fig:3}, \ref{fig:4} and \ref{fig:5}.

For our obstruction-based model, the sampled graphs are instances of 
obstruction graphs we generated according to the below described methods. 
In the non-obstruction model we take subgraphs of the test data 
(reproducing the results of Dai et al \cite{Dai18})

\subsection*{Vertex Cover Obstructions}
\label{Obstructions}

As mentioned in the previous section, designing an algorithm for computing all the elements of $\mathcal{O}b(k)$ is close to impossible.
We use the following two methods described in \cite{Dinneen07} to construct a sufficient number of connected obstructions. The reason behind our use of connected obstructions stems from the fact that every connected component of an obstruction is an obstruction for a smaller value of $k$ (so already computed at an earlier stage).

Given a connected graph $G \in \mathcal{O}b(k)$, we produce a graph $G' \in \mathcal{O}b(k+1)$ by applying Algorithms \ref{method1} or \ref{method2} described below.

\begin{algorithm}
\caption{Method1 \cite{Dinneen07}}
\label{method1}
\begin{algorithmic}
\STATE $V(G') \gets V(G) \cup \{v', v''\}$ //$v', v''$ new vertices
\STATE $E(G') \gets E(G) - \{v_1v_2\}$ \\
\STATE$E(G') \gets E(G') \cup \{v_1v', v'v'', v''v_2\} $\\
\end{algorithmic}
\end{algorithm}

\begin{algorithm}
\caption{Method2 \cite{Dinneen07}}
\label{method2}
\begin{algorithmic}
\STATE $V(G') \gets V(G) \cup \{v'\}$ //$v'$ is a new vertex\\
\STATE $E(G') \gets E(G) \cup \{v'v\} \cup \{v'u\mid u \in \mathcal{N}(v)\}$\\
\end{algorithmic}
\end{algorithm}

Our strategy for computing obstructions in this paper is based on using Algorithms \ref{method1} and \ref{method2}, starting from $\mathcal{O}b(3)$. 
The two methods are likely to produce many duplicates (or isomorphic) graphs. To reduce the number of produced obstructions, we used Nauty \& Traces C-library \cite{McKay201494}.
Consequently we generated the following numbers of connected obstructions reported in Table \ref{tab:obstructions} below. 

\begin{center}

\begin{table}[htb!]
    \centering
    \begin{tabular}{|c|c|c|}%
        \hline
        \bfseries $k$	& \bfseries Our count & \bfseries Exact count
            \begin{filecontents*}{obstructions.csv}
4,8,8
5,26,31
6,124,188
7,728,1930
8,5118,unknown
9,39900,unknown
10,335537,unknown
\end{filecontents*}
\csvreader[]{obstructions.csv}{}
        {\\\hline\csvcoli&\csvcolii&\csvcoliii}\\
        \hline
    \end{tabular}
    \caption{Number of obtained connected obstructions for size-$k$ vertex cover; exact counts obtained from \cite{Cattell94}, \cite{Dinneen2012} and
    \cite{Dinneen02}.}
    \label{tab:obstructions}
\end{table}

\end{center}

\section{Experiments}

To assess the utility of our obstructions-based training, we adopted the same neural network and experimental setup of Dai, et al in \cite{Dai18}. In fact we used the same hyper-parameters and setup, making sure we only change the training and validation data between the two approaches.

Training the network starts by randomly initializing its weights and at each iteration/epoch we pass the training data to the network to improve its weights and then evaluate the network on the validation data and finally take the best weights over all iterations (the weights that gave the best results on the validation data).

\begin{figure}[htb!]
    \centering
    \includegraphics[width=0.7\textwidth]{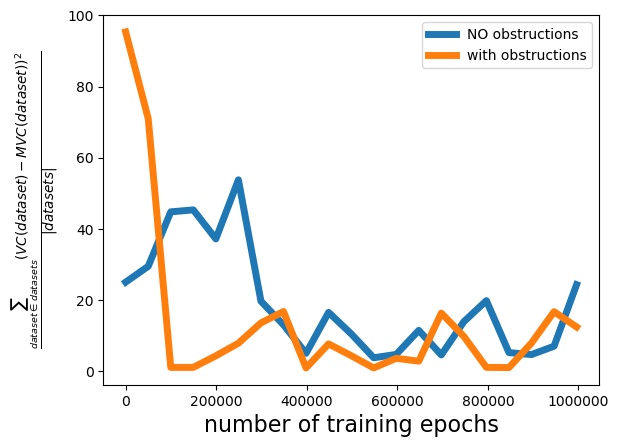}
    \caption{Average MSE of the two models (taken over all the data instances) after each epoch of training over their respective training sets.}
    \label{fig:1}
\end{figure}

As for the data sets, we used 19 widely used graphs from the Stanford Large Network Dataset Collection\footnote{https://snap.stanford.edu/data/} and networkrepository\footnote{http://networkrepository.com/index.php}
as well as the meme-tracker data used by \cite{Dai18}. We compared the size of the vertex cover produced by Dai's random subsets approach in \cite{Dai18} versus our obstructions' approach. Moreover, we compared both to the maximum-degree greedy heuristic and the factor-two approximation algorithm, dubbed Alg1 and Alg2, respectively.
We report our findings in Table \ref{tab:results}. As can be seen, the two approaches adopted by Dai et al. and our work are comparable. They both consistently outperform Alg2 and are nearly identical to Alg1. 
Despite the much smaller training set, the use of obstructions proved to be highly effective. In fact, 
we almost always outperform the random subsets approach of Dai et al., sometimes significantly as in the example of the MANN-a45 graph, and we obtain a vertex cover solver that is comparable to the best-known heuristic.
The main advantage of our approach is two-fold:

\begin{enumerate}
    \item Fast convergence (our method reached its best version in $\approx3000$ iterations while Dai's method used $\approx700,000$) 
    
    \item Stability of learning, as shown in Figure \ref{fig:1}
\end{enumerate}

\begin{table}[htb!]
\centering
    \resizebox{340pt}{!}{
\begin{tabular}{ | m{6.5em} | m{1cm} | m{1cm} | m{1cm} | m{1cm} | m{2cm}| m{2cm} | }
\hline
\bfseries Graph & \bfseries $\|V\|$ & \bfseries $\|E\|$ & \bfseries Alg1        & \bfseries Alg2        & 
\bfseries Random subsets      & \bfseries Obstructions method\\
\hline
C2000-5 & 2000 & 999164 & 1989 & 1996 & 1991 & 1991\\
\hline
C250-9 & 250 & 3141 & 216 & 236 & 214 & 211\\
\hline
C500-9 & 500 & 12418 & 453 & 484 & 456 & 453\\
\hline
MANN-a27 & 378 & 702 & 260 & 280 & 266 & 261\\
\hline
MANN-a45 & 1035 & 1980 & 704 & 740 & 740 & 705\\
\hline
brock800-1 & 800 & 112095 & 785 & 796 & 788 & 785\\
\hline
c-fat200-1 & 200 & 18366 & 187 & 198 & 188 & 192\\
\hline
c-fat200-5 & 200 & 11427 & 141 & 192 & 142 & 142\\
\hline
gen400-p0-9-55& 400 & 7980 & 370 & 386 & 371 & 371\\
\hline
gen400-p0-9-65& 400 & 7980 & 367 & 384 & 367 & 368\\
\hline
hamming10-2 & 1024 & 5120 & 511 & 948 & 513 & 515\\
\hline
hamming8-4 & 256 & 11776 & 239 & 252 & 240 & 240\\
\hline
keller4 & 171 & 5100 & 162 & 166 & 163 & 163\\
\hline
p-hat1500-1 & 1500 & 839327 & 1490 & 1498 & 1493 & 1494\\
\hline
p-hat700-2 & 700 & 122922 & 658 & 688 & 662 & 669\\
\hline
p-hat700-3 & 700 & 61640 & 642 & 684 & 647 & 642\\
\hline
san400-0-9-1 & 400 & 7980 & 349 & 386 & 350 & 350\\
\hline
sanr200-0-9 & 200 & 2037 & 161 & 184 & 163 & 164\\
\hline
sanr400-0-7 & 400 & 23931 & 382 & 394 & 386 & 384\\
\hline
Meme-tracker & 960 & 4888 & 482 & 658 & 480 & 481\\
\hline
\end{tabular}
}
\caption{Performance on real graphs: a comparison between the best performing network of \cite{Dai18} (after $\approx700,000$ epochs) vs. our model after 3000 epochs.}
    \label{tab:results}
\end{table}

We present in Figure \ref{fig:1} the mean squared error over time of each algorithm. In fact, after each training iteration we computed $\frac{1}{20} \sum_G (algorithm_i(G) - vc(G))^2$ where the algorithms are Alg1, Alg2, the model trained without obstructions (Dai et al \cite{Dai18}) and the same model trained with obstructions.

\begin{figure} [hbt!]
    \centering
    \includegraphics[width=0.75\textwidth]{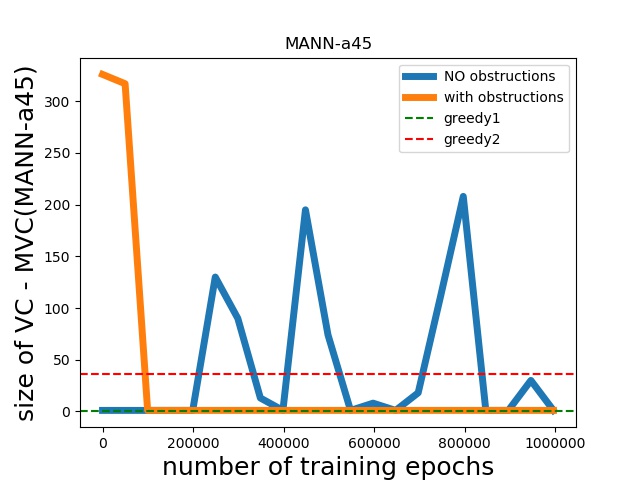}
    \caption{The output of the two models over the MANN-a45 dataset.}
    \label{fig:2}
\end{figure}

\begin{figure} [hbt!]
    \centering
    \includegraphics[width=0.75\textwidth]{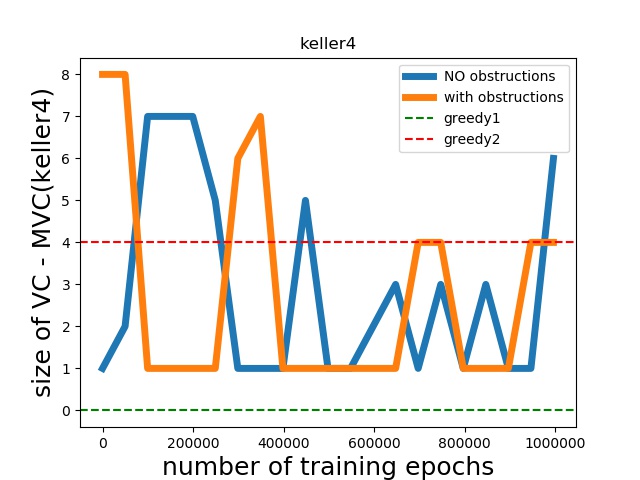}
    \caption{The output of two models over the keller4 dataset.} 
    \label{fig:3}
\end{figure}

\begin{figure} [hbt!]
    \centering
    \includegraphics[width=0.75\textwidth]{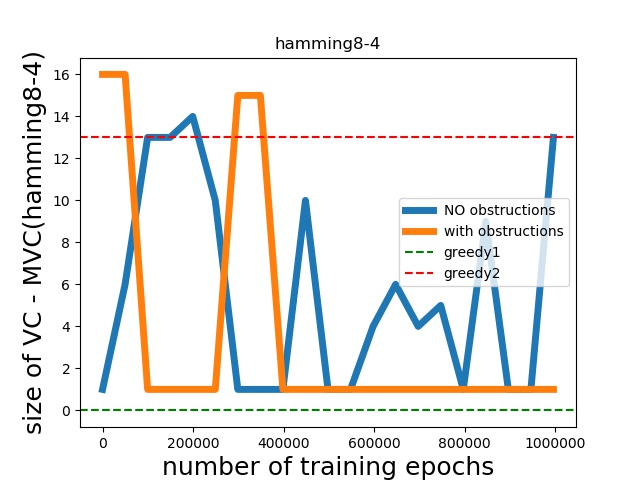}
    \caption{The output of the two models over the hamming8-4 dataset.} 
    \label{fig:4}
\end{figure}

\begin{figure} [htb!]
    \centering
    \includegraphics[width=0.75\textwidth]{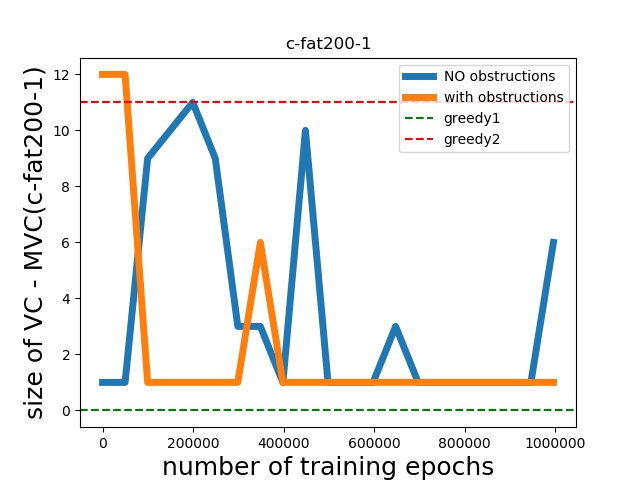}
    \caption{The output of the two models over the c-fat200-1 dataset.} 
    \label{fig:5}
\end{figure}

Finally, Figures \ref{fig:2}, \ref{fig:3}, \ref{fig:4} and \ref{fig:5} show the error over time for four of the 20 data sets, 
chosen so as to summarize the average overall performance.

\section{Conclusion}

In this paper we combined methods from graph theory and machine learning in an attempt to achieve improved approximation techniques for optimization problems. Our approach applies to what we call minor-closed problems such as Vertex Cover, Feedback Vertex Set, Treewidth, Cycle Packing, etc. The main idea is to use the notion of an obstruction for training.

The reported results prove that the use of obstructions has a tremendous impact on the training time, reducing the number of iterations from $\approx700K$ to $\approx3K$ in the training time needed for convergence.
Moreover, although we used only a small subset of vertex cover obstructions, we consistently obtained close to optimum solutions, and sometimes better than previously reported results, even at an early stage during training.

Finally, we have reported preliminary (proof of concept) results at this stage. Although our work thus far is restricted to computing the vertex cover obstruction sets and using them for training a particular network model, we believe this work will initiate a number of projects on various minor-closed problems, hopefully reviving the work on developing more techniques for computing obstruction sets.

\bibliographystyle{abbrv}
\bibliography{references}

\end{document}